\documentclass[fleqn]{elsarticle}
\usepackage{multirow}
\usepackage{amsmath}
\usepackage{amssymb}
\usepackage{amstext}
\usepackage{epsfig}
\usepackage{graphicx}
\usepackage{bm}
\usepackage{eufrak}
\usepackage{subfigure}
\usepackage{appendix}
\usepackage{color}
\usepackage{algorithm}
\usepackage{algorithmic}

\def\yhat{\hat{y}}

\newcommand{\appsection}[1]{\let\oldthesection\thesection
  \renewcommand{\thesection}{Appendix \oldthesection}
  \section{#1}\let\thesection\oldthesection}

\begin{document}

\begin{frontmatter}

\title{Short note on the behavior of recurrent neural network for noisy dynamical system}

\author{Kyongmin Yeo${}^*$}
\cortext[cor]{Corresponding author}
\ead{kyeo@us.ibm.com}

\address{IBM T.J. Watson Research Center, Yorktown Heights, NY, USA}

\begin{abstract}
The behavior of recurrent neural network for the data-driven simulation of noisy dynamical systems is studied by training a set of Long Short-Term Memory Networks (LSTM) on the Mackey-Glass time series with a wide range of noise level. It is found that, as the training noise becomes larger, LSTM learns to depend more on its autonomous dynamics than the noisy input data.  As a result, LSTM trained on noisy data becomes less susceptible to the perturbation in the data, but has a longer relaxation timescale. On the other hand, when trained on noiseless data, LSTM becomes extremely sensitive to a small perturbation, but is able to adjusts to the changes in the input data.
\end{abstract}


\end{frontmatter}

\section{Introduction}
Recently, deep learning has attracted a great attention for the modeling of physical systems due to its strength in the data-driven discovery of nonlinear dynamics \cite{Lusch18,Raissi19,Yeo19}. Among the deep learning techniques, recurrent neural network (RNN) provides a powerful tool to learn complex multiscale dynamics from the data \cite{Jaeger04,LeCun15}. Although RNN is shown to be very effective in modeling complex dynamics \cite{Yeo19,Pathak18,Vlachas18}, RNN is treated as a black-box model and the behavior of RNN itself is not well understood.

One of the distinguishing characteristics of RNN is its remarkable de-noising capability, as shown in figure \ref{fig:MG_one_step}. RNN makes a sequential update, where RNN makes a prediction for time ($t+1$) given the noisy observation at ($t$),  update its internal states, and repeat the process sequentially for time marching. It is shown that, even when the observation provided as an input is corrupted by a large noise, the prediction is always close to the ground truth. Considering that RNN consists of a set of deterministic functions, it is not clear how RNN can decompose the stochastic noise from the dynamics and make an accurate prediction.

Here, we study the behavior of RNN for noisy observations to provide an insight on the de-noising capability of RNN. The relative contributions of the noisy observations and the internal phase dynamics on the time evolution of RNN are investigated under a range of noise-to-signal ratio. It is found that, at high noise-to-signal ratio, RNN learns to suppress the contributions from the noisy observations, which makes the time evolution of RNN becomes more autonomous.

\section{Numerical Experiments}

\begin{figure}
  \centering
  \includegraphics[width=0.6\textwidth]{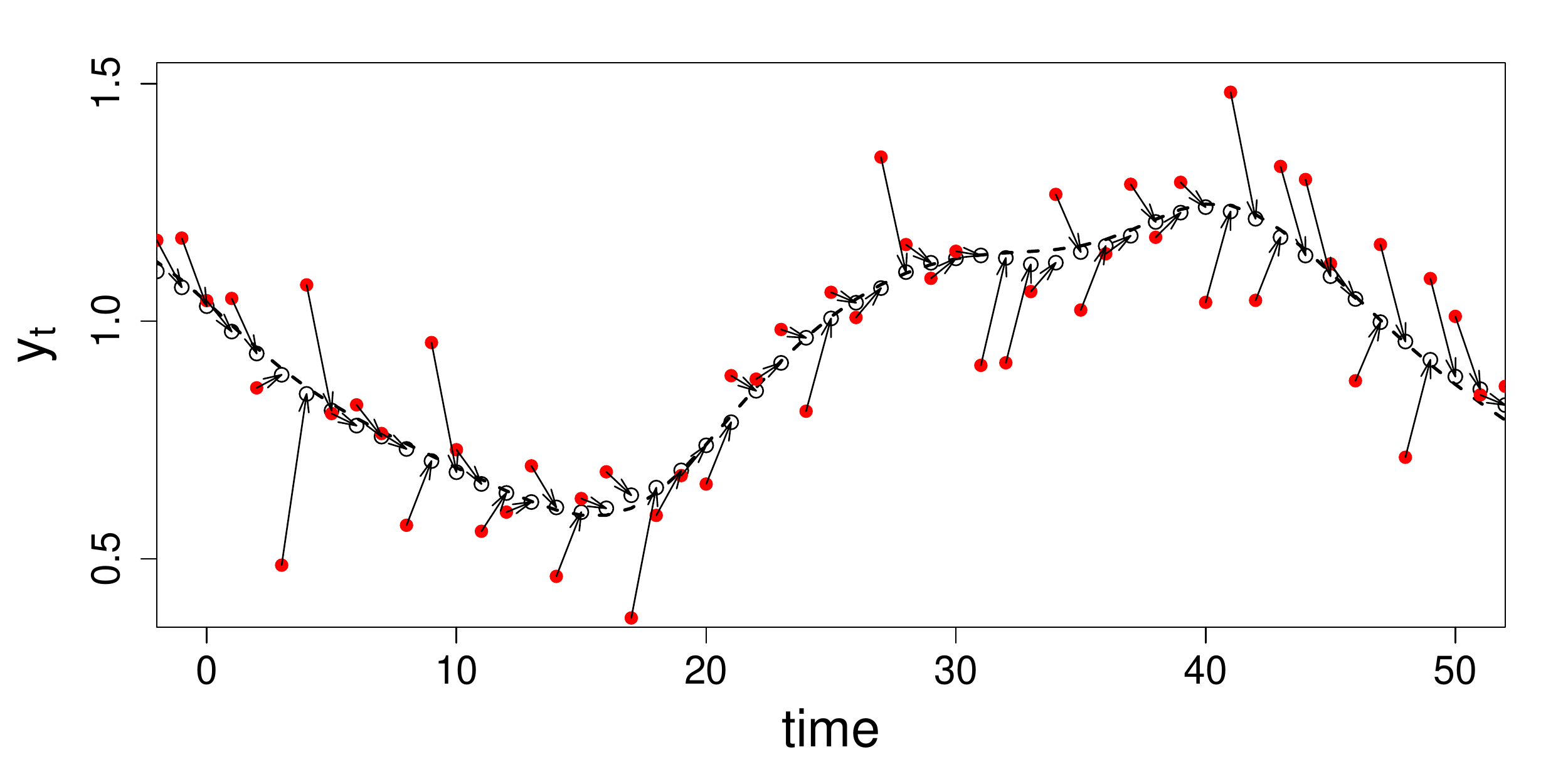}
  \caption{Sequential prediction of LSTM for a high noise level ($\sigma = 0.64$). The solid (${\color{red}\bullet}$) and hollow ($\circ$) symbols denote the noisy data and LSTM prediction, and the dashed line is the ground truth.} \label{fig:MG_one_step}
\end{figure}

Here, we use noisy observation of the Mackey-Glass time series \cite{Yeo19,Mackey77} to investigate the behavior of RNN. The Mackey-Glass equation is
\begin{equation} \label{eqn:MG}
\frac{d \mu(t)}{dt} = \frac{0.2 \mu(t-17)}{1+\mu^{10}(t-17)} - 0.1 \mu(t).
\end{equation}
The noisy data is generated by adding a white Gaussian noise to $\mu(t)$ with the sampling interval of $\delta t = 1$;
\begin{equation}
y_n = \mu(n\delta t) + \nu \epsilon_t,
\end{equation}
in which $\nu$ is the standard deviation of $\mu$ and $\epsilon_t \sim \mathcal{N}(0,\sigma^2 )$. Seven noise levels are considered, $\sigma \in (0, 0.02, 0.04, 0.08, 0.16,0.32,0.64)$.

The Long Short-Term Memory network (LSTM) \cite{Hochreiter97} is used for the recurrent neural network. The equations of LSTM used in this study are
\begin{align}
\text{input network:}& ~~\bm{z}_t = \tanh(\bm{W}_h \bm{h}_t + \bm{W}_y y_t), \label{eqn:net_in}\\
\text{gate functions:}&~~ \bm{G}_m = \bm{\varphi}( \bm{W}_m \bm{z}_t + \bm{b}_m),~~\text{for}~~~m\in(i,o,f) \label{eqn:gates}\\
\text{internal state:}&~~ \bm{s}_{t+1} = \bm{G}_f \odot \bm{s}_t + \bm{G}_i \odot \bm{\varphi}( \bm{W}_s\bm{z}_t + \bm{b}_s ), \label{eqn:cell_state}\\
\text{output state:}&~~ \bm{h}_{t+1} = \bm{G}_o \odot \tanh(\bm{s}_{t+1}), \label{eqn:cell_out}\\
\text{output network:}&~~ \yhat_{t+1} = \bm{W}_{y_2} \tanh(\bm{W}_{y_1} \bm{h}_{t+1} + \bm{b}_{y_1}) + b_{y_2}, \label{eqn:net_out}
\end{align}
in which $\bm{\varphi}$ is the Sigmoid function, $\odot$ is the Hadamard product, $\bm{s}_t \in \mathbb{R}^N$ and $\bm{h}_t \in \mathbb{R}^N$ are the internal states, $N$ is the number of memory cells, and $\yhat_t$ is the LSTM prediction. The weight matrices and bias vectors are $\bm{W}_m \in \mathbb{R}^{N\times N}$ and $\bm{b}_m \in \mathbb{R}^N$ for $m \in (h,i,o,f,s,y_1)$, and $\bm{W}_y \in \mathbb{R}^{N\times 1}$ and $\bm{W}_{y_2} \in \mathbb{R}^{1\times N}$. The size of LSTM is $N=128$. The LSTM is trained to minimize the mean-square loss,
\[
\mathcal{L} = \sum_{t=1}^T \frac{1}{2}(y_t  - \yhat_t)^2,
\]
by using a variant of the stochastic gradient descent method, called ADAM\cite{Kingma15}. Total seven LSTMs are trained, one for each noise level. In the training, the data is normalized by the minimum and maximum of each time series, so that the input data is $y^*_t \in [-0.5,0.5]$. The model output is scaled back to the original scale. Here, we are interested in the sequential problem,
\begin{equation}
\yhat_{t+1} = \Psi_\sigma(y_t,\bm{s}_t,\bm{h}_t),
\end{equation}
in which $\Psi_\sigma$ denotes the LSTM trained against a noise level $\sigma$, e.g., $\Psi_0$ or $\Psi_{0.16}$. Figure \ref{fig:MG_one_step} shows the sequential prediction of LSTM for $\sigma = 0.64$.

\begin{figure}
  \centering
  \includegraphics[width=0.48\textwidth]{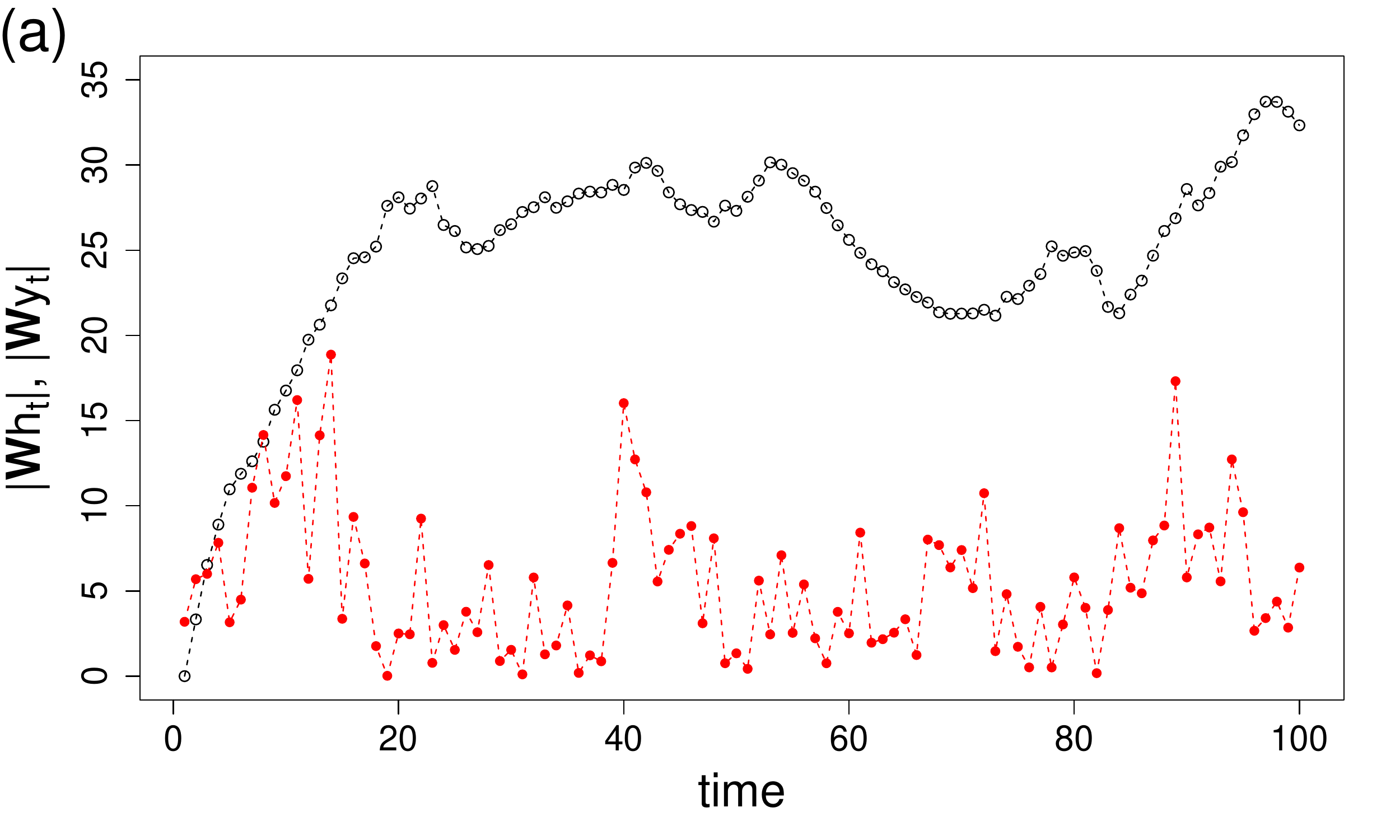}
  \includegraphics[width=0.48\textwidth]{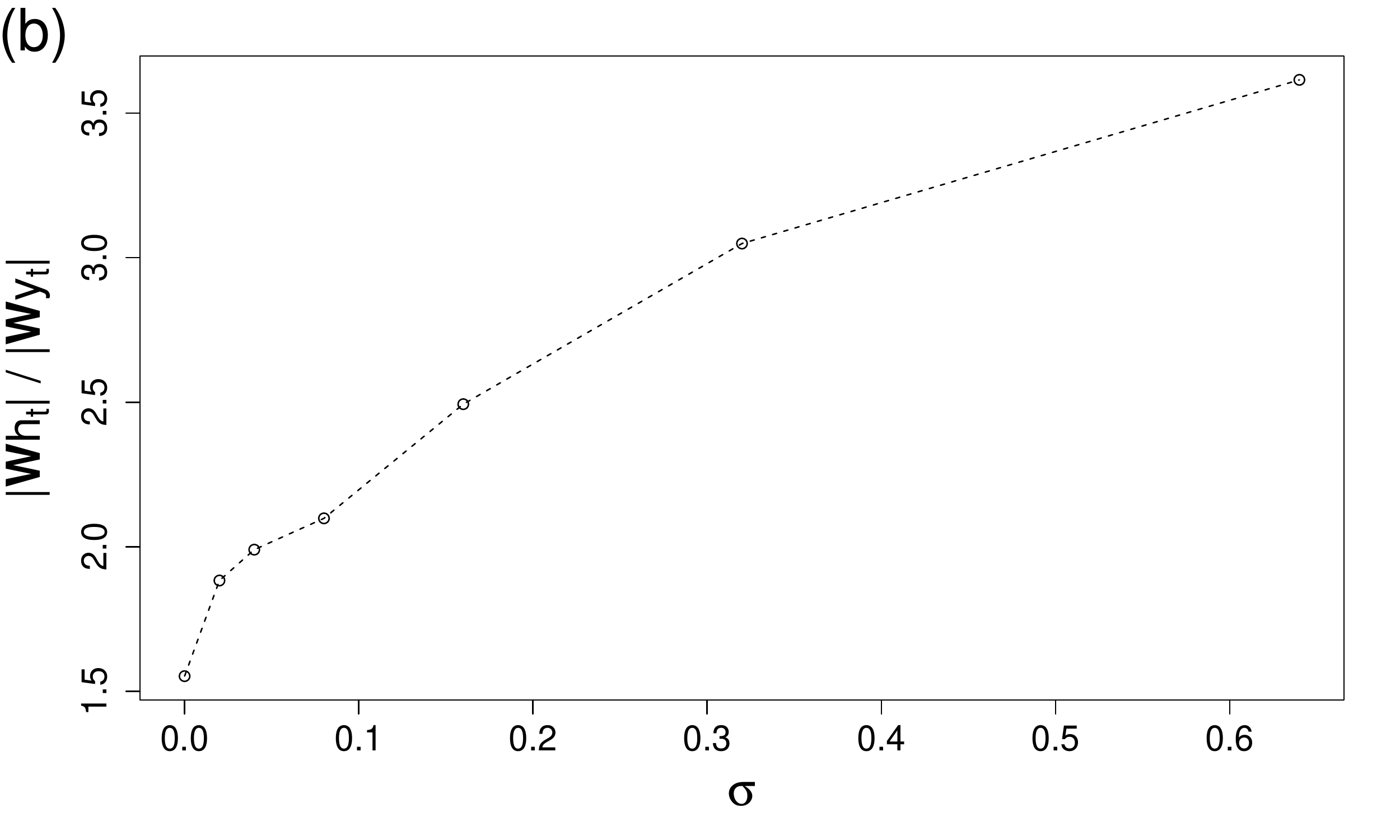}
  \caption{ (a) Time evolution of $\|\bm{W}_h \bm{h}_t\|_1$  ($\circ$) and  $\|\bm{W}_y y_t\|_1$ (${\color{red}\bullet}$) for $\sigma = 0.64$.  (b) Estimated ratio between $\|\bm{W}_h \bm{h}_t\|_1$ and $\|\bm{W}_y y_t\|_1$ in terms of the noise level, $\sigma$.  } \label{fig:ratio_internal}
\end{figure}

Equations (\ref{eqn:net_in}--\ref{eqn:cell_state}) indicate that the temporal evolution of the internal dynamics, $\bm{s}_t$, is fully dictated by the relative contributions of $\bm{h}_t$ and $y_t$ to the input signal, $\bm{z}_t$. Figure \ref{fig:ratio_internal} (a) shows the time evolution of $\|\bm{W}_h\bm{h}_t\|_1$ and $\|\bm{W}_yy_t\|_1$ for $\sigma = 0.64$. The trajectory of  $\|\bm{W}_h\bm{h}_t\|_1$ shows that the internal state of LSTM slowly relaxes toward a stationary state from an initial condition, $\bm{h}_1 = \bm{0}$. Even when the data is very noisy, as reflected in  $\|\bm{W}_y y_t\|_1$, the trajectory of  $\|\bm{W}_h\bm{h}_t\|_1$ remains relatively smooth.

The relative contribution between  $\bm{h}_t$ and $y_t$ to $\bm{z}_t$ is shown in figure \ref{fig:ratio_internal} (b). The relative contribution is estimated as
\begin{equation}
\alpha = \frac{1}{T \times N} \sum_{t=1}^T\sum_{i=1}^N \frac{ |(\bm{W}_h \bm{h}_t)_i |}{|(\bm{W}_y y_t)_i | + |(\bm{W}_h \bm{h}_t)_i |} ~~\Rightarrow~~ \frac{|\bm{W}_h \bm{h}_t|}{|\bm{W}_y y_t|} \simeq  \frac{\alpha}{1-\alpha},
\end{equation}
in which $(\bm{a})_i$ denotes the $i$-th element of a vector $\bm{a}$. It is shown that, at higher noise level, the contribution of $\bm{h}_t$ overwhelms that of $y_t$. 
The results suggest that, as the data becomes noisier, LSTM becomes less sensitive to the data and depends more on its own dynamics.

\begin{figure}
  \centering
  \includegraphics[width=0.6\textwidth]{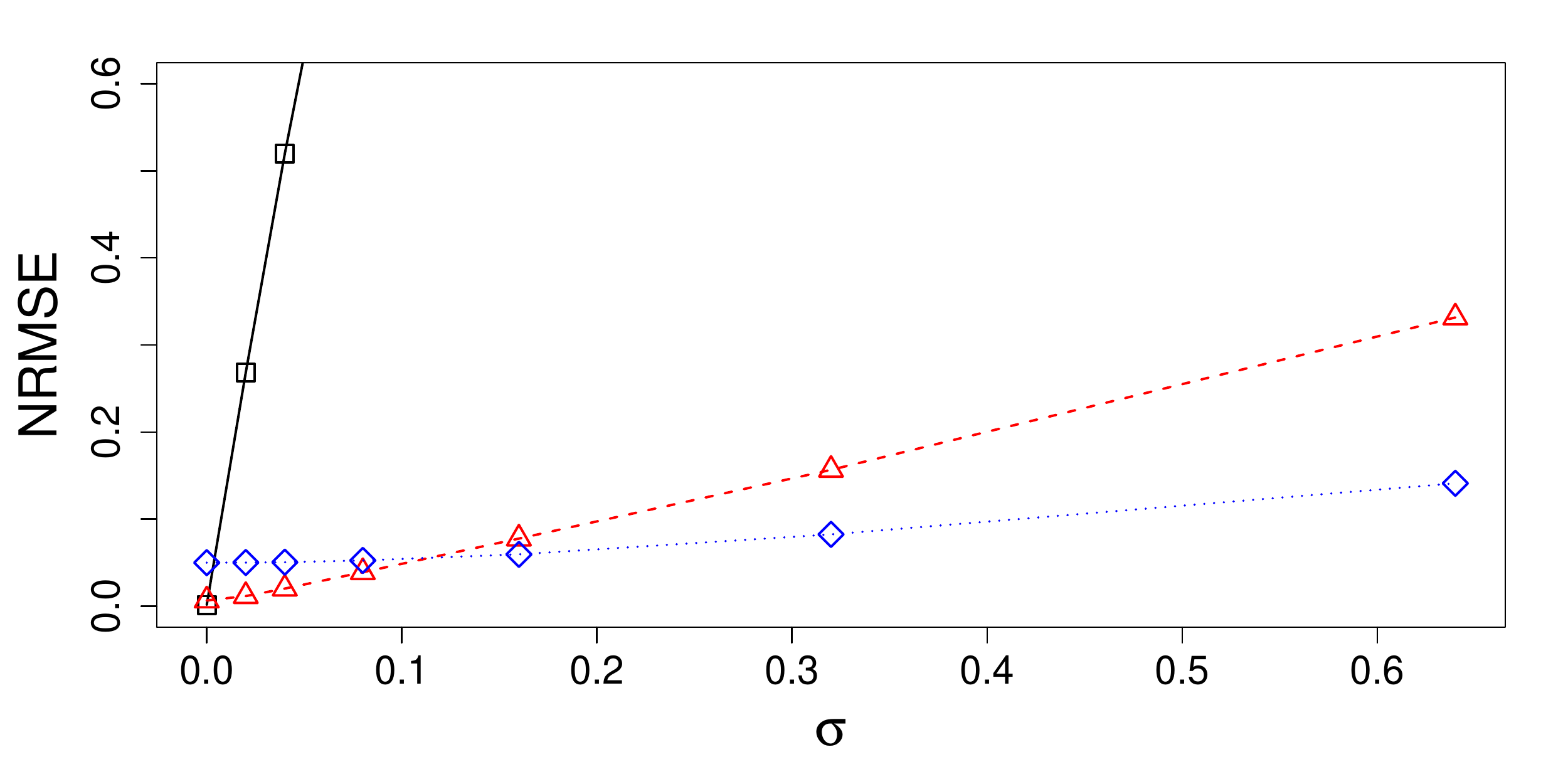}
  \caption{ NRMSE in terms of the noise level ($\sigma$) for LSTM models trained against $\sigma = 0$ ($\Box$), 0.02 (${\color{red}\triangle}$), and 0.64 (${\color{blue}\diamond}$). } \label{fig:comp_error}
\end{figure}

Figure \ref{fig:comp_error} shows the normalized root mean-square errors (NRMSE) of the sequential predictions as a function of the noise level for three LSTMs trained against $\sigma = 0, 0.02, 0.64$, i.e., $\Psi_0$, $\Psi_{0.02},$ and $\Psi_{0.64}$. NRMSE is defined as
\[
e_\mu = \frac{1}{\nu} \left[\frac{1}{T}\sum_{t=1}^T (\mu(t\delta t) - \yhat_t)^2 \right]^{1/2}.
\]
It is shown that $\Psi_{0}$ is extremely sensitive to the noise. For the smallest noise, $\sigma = 0.02$, $e_\mu$ of the zeroth-order prediction, $\yhat_{t+1} = y_t$, is 0.147, while $\Psi_0$ has $e_\mu$ of 0.268. When LSTM is trained even with a small noise, the sensitivity to the noise in the data becomes much smaller. For $\Psi_{0.02}$, $e_\mu = 0.332$ at $\sigma = 0.64$, which is only about half of the zeroth-order prediction NRMSE ($e_\mu = 0.658$). As expected, $\Psi_{0.64}$ is much less sensitive to the noise. The error changes from $e_\mu = 0.050$ at $\sigma = 0$ to $e_\mu = 0.141$ at $\sigma=0.64$. However, because $\Psi_{0.64}$ relies more on the internal dynamics, the accuracy is not improved even when the data becomes noiseless. There is virtually no change in $e_\mu$ of $\Psi_{0.64}$ for $\sigma = 0 \sim 0.04$.

\begin{figure}
  \centering
  \includegraphics[width=0.32\textwidth]{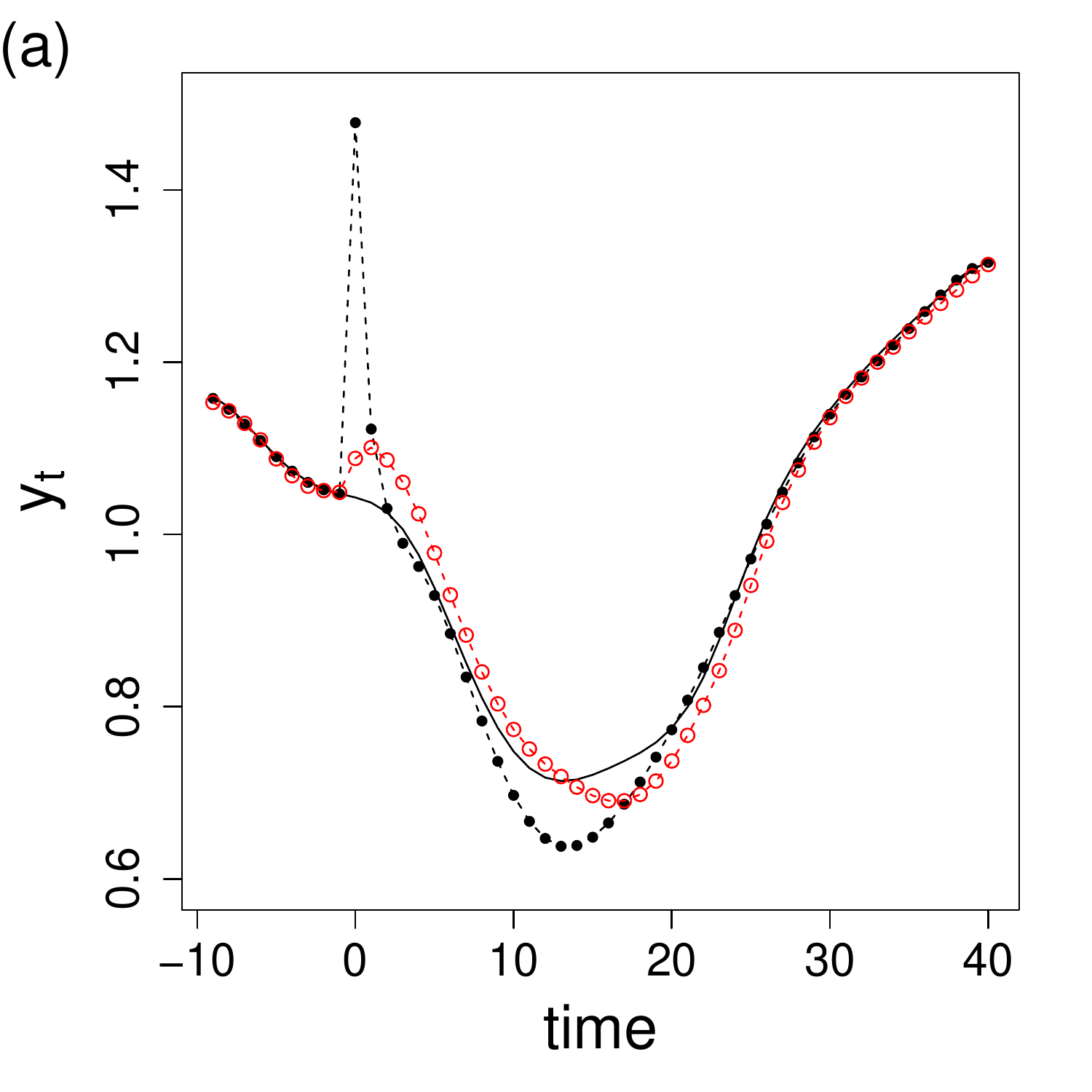}
  \includegraphics[width=0.32\textwidth]{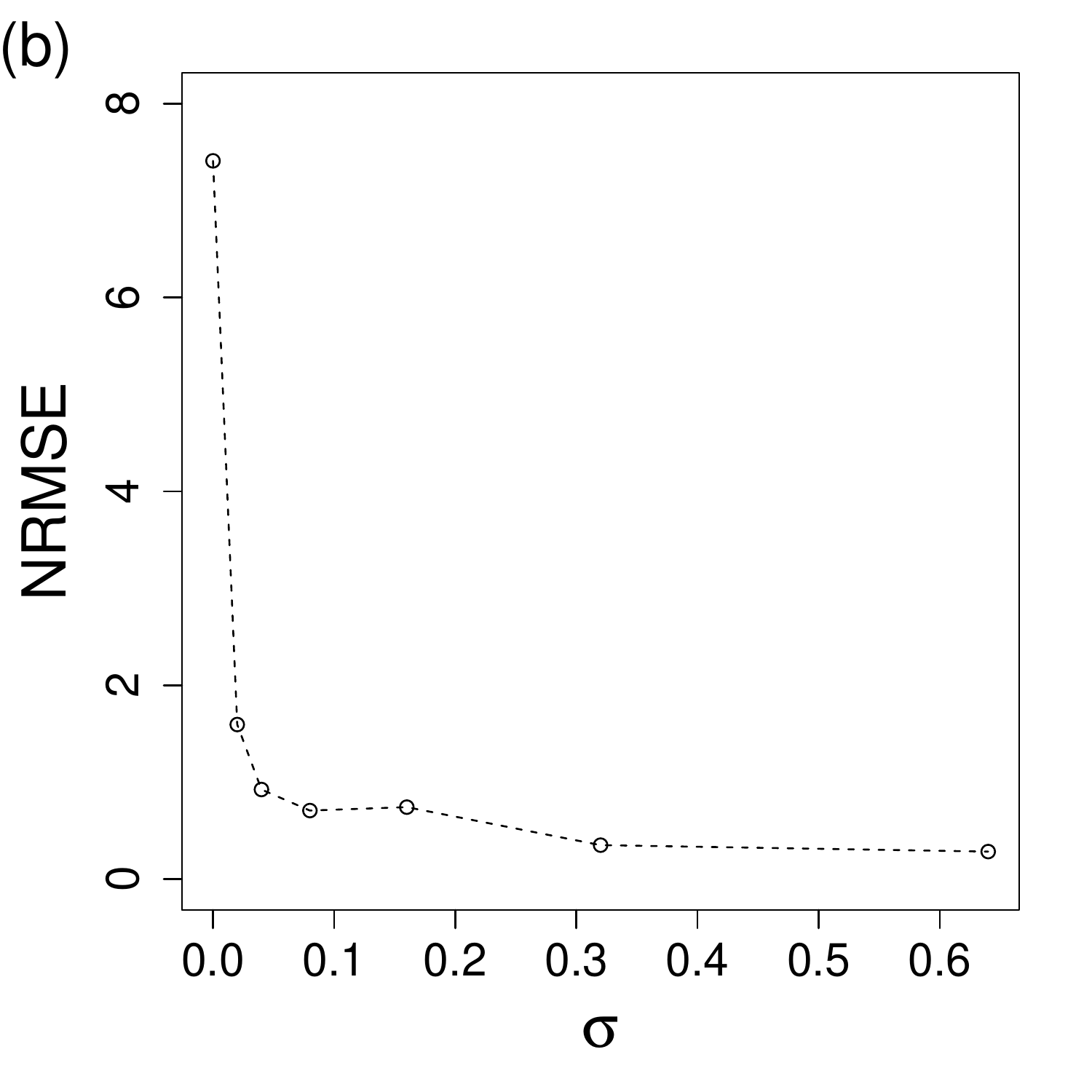}
  \includegraphics[width=0.32\textwidth]{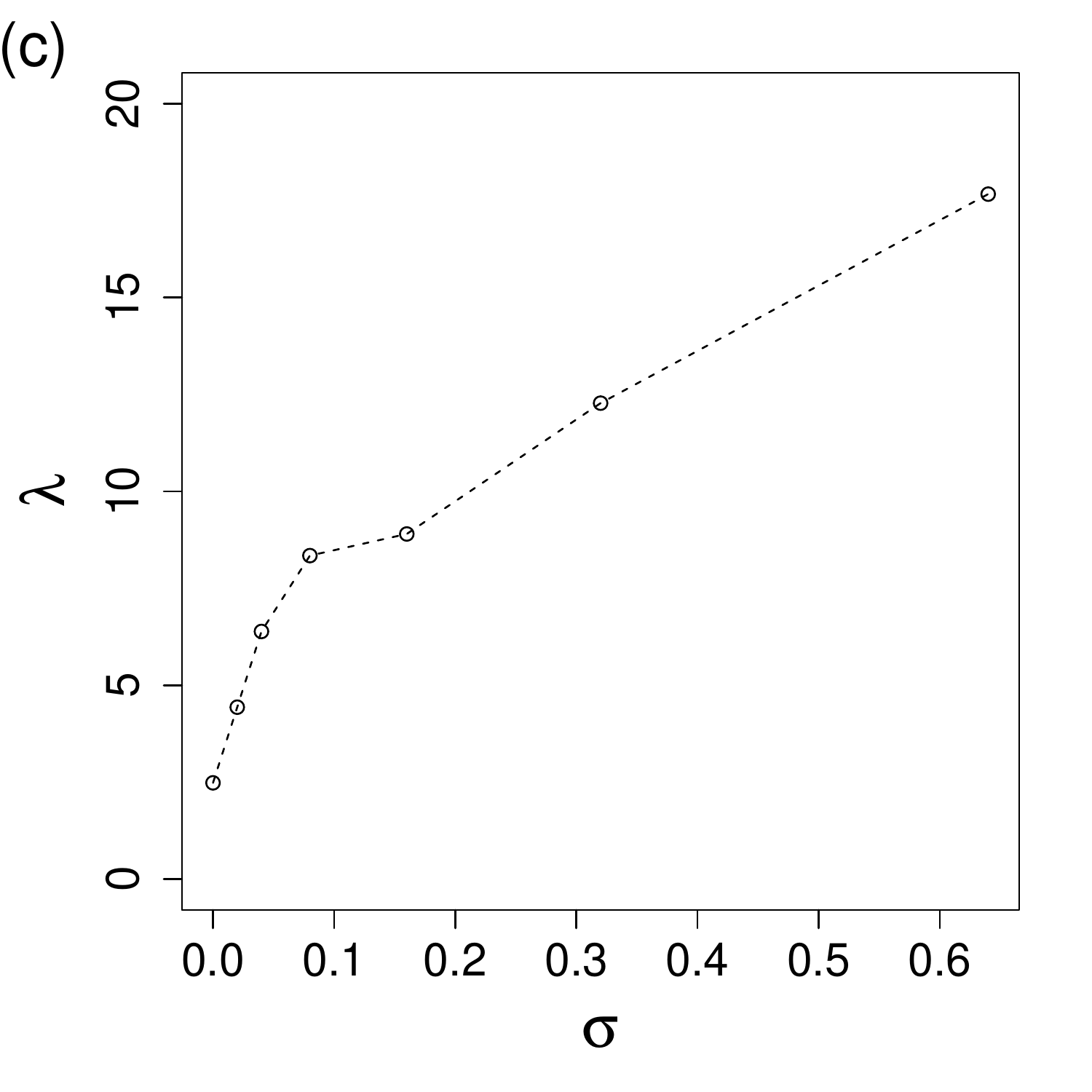}  
  \caption{ (a) Impulse response of $\Psi_{0.02}$ ($\bullet$) and $\Psi_{0.64}$ (${\color{red}\circ}$) with the ground truth, $\mu(t)$ ($\frac{~}{~}\,\frac{~}{~}\,\frac{~}{~}$). (b) Initial errors of $\Psi_\sigma$ at the impulse. (c) Relaxation timescale of $\Psi_\sigma$.} \label{fig:relax_test}
\end{figure}

The relaxation of $\Psi$ after an impulse is shown in figure \ref{fig:relax_test}. The simulations are performed for the noiseless data, i.e., $\sigma = 0$. At every 150$\delta t$, an impulse is added to the input data such that
\[
\yhat_{t+1} = \Psi(y_t + 1,\bm{s}_t,\bm{h}_t),
\]
and, the noiseless data is provided to LSTM for the next 149$\delta t$. Figure \ref{fig:relax_test} (a) shows the impulse responses of $\Psi_{0.02}$ and $\Psi_{0.64}$. $\Psi_{0.02}$ exhibits a large deviation from $\mu(t)$ right after the impulse, then converges to $\mu(t)$ around $t = 20 \delta t$. On the other hand, for $\Psi_{0.64}$, although the magnitude of the deviation is much smaller, it takes longer for the effects of the impulse to vanish, around $t = 25 \sim 30 \delta t$. 

The average deviation from $\mu(t)$ is computed as NRMSE,
\[
e_n = \frac{1}{\nu}\langle (\mu(n\delta t)-\yhat_n)^2 \rangle^{1/2},~~\text{for}~~n = 0,\cdots,149,
\]
in which $\langle \cdot \rangle$ denotes an ensemble average. Figure \ref{fig:relax_test} shows the initial deviation due to the impulse, $e_0$. In general, LSTM trained against smaller $\sigma$ shows larger $e_0$. The initial deviation of $\Psi_0$ is about 1.7 times larger than the magnitude of the impulse, which is reduced to 0.36 of the impulse for $\Psi_{0.02}$.

A relaxation timescale is estimated as
\begin{equation}
\lambda = \sum_{t=0}^{149} \frac{ e_t  - e_\mu}{ e_0 - e_\mu} \delta t \simeq \int^T_0 \frac{e(t) - e_{\mu}}{e_0-e_{\mu}} dt.
\end{equation}
Note that $e_t = e_{\mu}$ for $t$ large. The relaxation timescale is shown in figure \ref{fig:relax_test} (c). It is shown that the relaxation timescale increases with the training noise, $\sigma$. Because $\Psi_\sigma$ trained for a larger $\sigma$ has a larger inertia, even though the initial deviation is smaller, it takes longer for the impulse effect to vanish. On the other hand, $\Psi_\sigma$ for a smaller $\sigma$ quickly adjusts the dynamics after the large initial deviation, which makes $\lambda$ smaller.

\section{Conclusions} \label{sec:conclusions}

The behavior of RNN for noisy observations of a nonlinear dynamical system is numerically investigated by using LSTM on the noisy Mackey-Glass time series. It is found that, when trained on a noisy data, LSTM learns to reduce the contributions of the noisy input data and spontaneously develops its own dynamics. As a result, when the noise in the training data becomes larger, LSTM becomes less sensitive to the input data and more reliance on its own internal dynamics. It is shown that LSTM trained on a larger noise is less susceptible to an impulse, but the effects of the impulse persists longer.  On the other hand, LSTM trained on noiseless data becomes extremely sensitive to a small perturbation in the input data. The results support the current heuristic in deep learning that injecting noise makes the model more robust.

\bibliographystyle{elsarticle-num}
\bibliography{LSTM_ref}

\end{document}